\documentclass[conference]{IEEEtran}
\IEEEoverridecommandlockouts
\usepackage{cite}
\usepackage{amsmath,amssymb,amsfonts}
\usepackage{algorithmic}
\usepackage{graphicx}
\usepackage{textcomp}
\usepackage{arabtex}
\usepackage{utf8}
\setcode{utf8}
\usepackage{csquotes} 
\MakeOuterQuote{"}
\usepackage{booktabs}
\usepackage{hyperref}
\usepackage[margin=0.75in]{geometry}

\usepackage{xcolor}

\makeatletter
\newcommand{\linebreakand}{%
  \end{@IEEEauthorhalign}
  \hfill\mbox{}\par
  \mbox{}\hfill\begin{@IEEEauthorhalign}
}
\makeatother
\begin{document}

\title{Improving Natural Language Inference in Arabic using Transformer Models and Linguistically Informed Pre-Training
}

\author{\IEEEauthorblockN{1\textsuperscript{st} Mohammad Majd Saad Al Deen\textsuperscript{\textsection}}
\IEEEauthorblockA{\textit{Fraunhofer IAIS and Hochschule Bonn-Rhein-Sieg} \\
Sankt Augustin, Germany}
\and
\IEEEauthorblockN{2\textsuperscript{nd} Maren Pielka\textsuperscript{\textsection}}
\IEEEauthorblockA{\textit{Fraunhofer IAIS} \\
Sankt Augustin, Germany \\
maren.pielka@iais.fraunhofer.de}
\and
\IEEEauthorblockN{3\textsuperscript{rd} Jörn Hees}
\IEEEauthorblockA{\textit{Hochschule Bonn-Rhein-Sieg} \\
Sankt Augustin, Germany}
\linebreakand
\IEEEauthorblockN{4\textsuperscript{th} Bouthaina Soulef Abdou}
\IEEEauthorblockA{\textit{Fraunhofer IAIS and University of Bonn} \\
Sankt Augustin, Germany}
\and
\IEEEauthorblockN{5\textsuperscript{th} Rafet Sifa}
\IEEEauthorblockA{\textit{Fraunhofer IAIS and University of Bonn} \\
Sankt Augustin, Germany}
}
\maketitle
\begingroup\renewcommand\thefootnote{\textsection}
\footnotetext{Both authors contributed equally.}
\endgroup

\begin{abstract}
This paper addresses the classification of Arabic text data in the field of Natural Language Processing (NLP), with a particular focus on Natural Language Inference (NLI) and Contradiction Detection (CD). Arabic is considered a resource-poor language, meaning that there are few data sets available, which leads to limited availability of NLP methods. To overcome this limitation, we create a dedicated data set from publicly available resources. Subsequently, transformer-based machine learning models are being trained and evaluated. We find that a language-specific model (AraBERT) performs competitively with state-of-the-art multilingual approaches, when we apply linguistically informed pre-training methods such as Named Entity Recognition (NER). To our knowledge, this is the first large-scale evaluation for this task in Arabic, as well as the first application of multi-task pre-training in this context.
\end{abstract}

\section{Introduction}

Natural Language Processing (NLP) in Arabic, also known as Arabic NLP, is a subfield of Artificial Intelligence (AI) that focuses on processing and analysing textual data in the Arabic language. It encompasses various technologies and methods for automating tasks such as text classification, sentiment analysis, and machine translation. The goal is to teach computers to understand and process Arabic language, enabling a range of applications including chatbots, text mining tools, and translation services. This is a challenging field due to the limited availability of training data and pre-trained models for the Arabic language.

Natural Language Inference (NLI), also referred to as Recognizing Textual Entailment (RTE), is a subfield of Natural Language Processing (NLP) that aims to determine the semantic relationship between two pieces of text, known as the "premise" and the "hypothesis". The task, as described by MacCartney et al. \cite{Natural_Logic_for_Textual_Inference}, involves identifying possible connections such as "entailment" (if the premise is true, the hypothesis must also be true), "contradiction" (if the premise is true, the hypothesis cannot be true, and vice versa), or "neutral" (there is no logical relationship between the two sentences; both can be true or false simultaneously). Accomplishing this task requires a machine learning algorithm to comprehend the semantics of a text, which poses a particularly challenging problem.

In this work, we apply existing deep learning approaches, namely AraBERT and XLM-RoBERTa, to the task of Natural Language Inference and Contradiction Detection in Arabic. To our best knowledge, this has not been attempted before in such a comprehensive manner, as previous studies were limited to a smaller number of data sources and models. We also employ an informed language modeling approach, by further pre-training the transformer model on an NER task, before fine-tuning it on the downstream tasks. This is a novel direction of research with respect to Arabic text. In addition, we introduce a new data collection for NLI/CD in Arabic language, which is publicly available on Github\footnote{\url{https://github.com/fraunhofer-iais/arabic_nlp/}}.

\section*{Related Work}
In the field of Natural Language Inference (NLI) and considering its impact on Question Answering (QA) tasks, Mishra et al. \cite{Mishra} conducted a research study. They utilized the RACE dataset \cite{G. Lai}, which is a large-scale reading comprehension dataset consisting of questions and answers from English exams for Chinese students. The authors converted a subset of the RACE dataset (containing 48,890 training examples, 2,496 validation examples, and 2,571 test examples) into an NLI format and compared the performance of a state-of-the-art model, RoBERTa [23], in both formats. To convert a reading comprehension question into a Natural Language Inference (NLI) format, the question was used as the premise, and each answer option was paraphrased as individual hypotheses. The same model architecture, comprising a RoBERTa encoder and a two-layer feed-forward network as the classification head, was employed for both QA and NLI. The results showed that the NLI model outperformed the QA model on the subset of the RACE dataset. This can be attributed to the more natural form of the hypotheses in the NLI model compared to the combination of question and answer option in the QA model.

With respect to NLI in Arabic, Jallad et al. \cite{K. A. Jallad} conducted a similar study and created their own dataset called arNLI, which consists of over 6,000 data points. The data was obtained by using machine translation from two English sources, namely SICK\footnote{\url{https://alt.qcri.org/semeval2014/task1/}} and PHEME\footnote{\url{https://www.kaggle.com/datasets/usharengaraju/pheme-dataset}}. The authors developed a system with three main components: text preprocessing (cleaning, tokenization, and stemming), feature extraction (contradiction feature vector and language model vectors), and a machine learning classification model. The morphological units were processed using the Snowball Stemmer (Porter2) algorithm. Various types of features were employed for feature extraction, including features for named entities, similarity features, specific stopword features, number, date, and time features, which were processed using different language models such as TF-IDF \cite{TFIDF} and Word2Vec embeddings \cite{Word2Vec}. To determine the relationship type between two sentences (contradiction, entailment, or neutral), various traditional machine learning classifiers were used, including Support Vector Machine (SVM), Stochastic Gradient Descent (SGD), Decision Tree (DT), ADA Boost Classifier, K-Nearest Neighbors (KNN), and Random Forest. The proposed solution by the authors was trained and evaluated on their own dataset (arNLI), and they reported that Random Forest achieved the best results on the arNLI dataset with an accuracy of 75\%.

Contradiction detection has received relatively limited attention in the literature compared to other tasks. De Marneffe et al. \cite{Marneffe} define a contradiction as a conflict between two statements that mutually negate each other. In a stricter logical sense, there is no possible world in which both statements can be simultaneously true. A looser definition, aligning better with human intuition, suggests that a contradiction exists when it is highly improbable for two statements to be simultaneously true. To identify these contradictions, the authors employ an approach based on syntactic analysis and semantic understanding of the text. They utilize syntactic parsing tools to establish the logical structure of the text, and semantic features such as antonymy, polarity, and numerical deviation to comprehend the meaning of words and sentences in the text. The authors note that detecting contradictions may be more challenging than recognizing entailment and requires a deep semantic understanding, possibly augmented by world knowledge.

Pucknat et al. \cite{L. Pucknat} conducted a study comparing the performance of four state-of-the-art models in NLI, particularly for contradiction detection, on German text data. These models were evaluated based on their performance on a machine-translated version of the well-known Stanford Natural Language Inference dataset (SNLI) \cite{SNLI} and the German test set of the Cross-Lingual NLI Corpus (XNLI) \cite{XNLI}. One key focus was to determine if the models were robust with respect to data selection and could potentially be applied in real-world scenarios. The XLM-RoBERTa model significantly outperformed the other models, likely due to its extensive pre-training and multi-head attention. However, the models did not generalize well to the XNLI data, indicating that the training corpus was limited in terms of topics and types of contradictions. The authors report an accuracy of 86.5\% when testing XLM-RoBERTa on the XNLI dataset.

Another methodology by Pielka et al. \cite{M. Pielka} focused on pre-training methods to integrate syntactic and semantic information into state-of-the-art model architectures. The authors presented a linguistically enhanced approach for pre-training transformer models. They incorporated additional knowledge about part-of-speech tags, syntactic analysis, and semantic relationships between words into the model. Their goal was to become independent of massive pre-training data resources by integrating as much external knowledge as possible into the model. Their approach was evaluated on the SNLI dataset, and they demonstrated that the smaller BERT model can be competitive with XLM-RoBERTa when enhanced with additional knowledge during pre-training. Their approach did not require additional data for pre-training, as it was trained on additional tasks using the same dataset that would later be used for fine-tuning. 

\section*{Data}
For this study, a self-constructed corpus was used, comprising data from three different sources. 
\begin{itemize}

 \item XNLI (Cross-Lingual NLI Corpus) \cite{XNLI}:
    The Arabic-translated section of the XNLI dataset was included as a source for the corpus. XNLI is a well-known benchmark dataset for cross-lingual natural language inference tasks, containing 7500 text pairs in 15 languages.

 \item SNLI (Stanford Natural Language Inference Corpus) \cite{SNLI}:
    The Arabic-translated section of the SNLI dataset by \cite{ArabicSNLI}, comprising 1332 manually translated sentence pairs, was also incorporated into the corpus. SNLI is a widely used dataset in English language for NLI, consisting of sentence pairs labeled with three relationship types: entailment, contradiction, and neutral. 

 \item arNLI (Arabic Natural Language Inference) \cite{K. A. Jallad}:
    The arNLI dataset, specifically created for the NLI task in Arabic language, was an additional source of data. This dataset consists of 6366 data points and was obtained through machine translation from two English sources.
   
\end{itemize}

Some examples from the data set are displayed in \autoref{beispiel_korpus}.

\begin{table*}[ht]
    \centering
    \resizebox{\textwidth}{!}{
    \begin{tabular}{ccccc} \toprule
             \Large\textbf{premise} & \Large\textbf{hypothesis} &  \Large\textbf{label} \vspace{1mm}  &  \vspace{1mm} \Large\textbf{premise\_en}   & \Large\textbf{hypothesis\_en}  \\ \midrule
        
        \begin{minipage}{4cm} \vspace{1mm} \textbf{}  \<لقد دخل أول فريق للتدخل\\
        السريع التابع لشرطة نيويورك
       \\ ردهة الشارع الغربي للبرج 
       \\ الشمالي واهم مستعدون لبدء
      \\التسلق حوالي ٥١:٩ صباحا>
        
        \textbf{}  \end{minipage}  & \begin{minipage}{4cm} \vspace{1mm}   \textbf{} \<\\\\كان البرج لا يزال قائماً \\
      في الساعة ٥١:٩ صباحا
      \\
      \\
      \\
      >
         \end{minipage}  &  0  & \begin{minipage}{4cm} \vspace{1mm}  
       \textbf{The first NYPD ESU team entered the West Street-level lobby of the North Tower and prepared to begin climbing at about 9:15 A.M}\\
        \end{minipage} & \begin{minipage}{4cm} \vspace{3mm}  
      \textbf{The tower was still standing at 9:15 AM.}\\
      \\
      \\
        \end{minipage} \\ \midrule
        \begin{minipage}{4cm} \vspace{1mm}  \<يتطلب الأمرشراكة بين 
        \\الدعم الخاص والتمويل
        \\الجامعي لمدرستنا القانونية 
        \\لمواصلة النمو في المكانة والتأثي
     \\>
        \end{minipage}
          & \begin{minipage}{4cm} \vspace{1mm}   \textbf{
        \<\\مدرسة القانون لدينا مدعومة\\
        جزئياً بواسطة مؤسسة ميلندا\\
         وبيل جيت
        \\
        \\>
        }
      \end{minipage}  & 1 & \begin{minipage}{4cm} \vspace{1mm} \textbf{It takes a partnership of private support and University funding for our law school to continue to grow in stature and influence \\
      }  \end{minipage} & \begin{minipage}{4cm} \vspace{2mm} \textbf{Our law school is supported in part by the Melinda and Bill Gates Foundation \\
      \\}  \end{minipage} \\ \midrule
        
               \begin{minipage}{4cm} \vspace{1mm}  \<يجب على الأمريكيين أيضًا\\
               أن يفكروا في كيفية القيام\\
               بذلك وتنظيم حكومتهم بطريقة\\
               مختلفة>
        \end{minipage}
          & \begin{minipage}{4cm} \vspace{1mm}   \textbf{
        \<\\يمكن تنظيم الحكومة فقط
        \\بطريقة واحدة وأي محاولة
      \\لتغييرها ستكون غبية
      \\
      \\>     }
      \end{minipage}  & 2 & \begin{minipage}{4cm} \vspace{1mm} \textbf{Americans should also consider how to do it-organizing their government in a different way. \\
      }  \end{minipage} & \begin{minipage}{4cm} \vspace{1mm} \textbf{The government can only be organized in one way and any attempt to change it would be foolish \\}  \end{minipage} \\ \midrule

       \begin{minipage}{4cm} \vspace{1mm} 
       \<البوابة التي تمثل جزءًا من\\
               جدار المدينة، لم يكن المقصود\\
               منها من البروسيين الأكثر\\
               براغماتية أكثر من مجرد قوس\\
               النصر كمعبر لفرض الرسوم>
        \end{minipage}
          & \begin{minipage}{4cm} \vspace{1.5mm}   \textbf{
        \<كانت البوابة مجرد قوس نصر.
        \\>     }
      \end{minipage}  & 2 & \begin{minipage}{4cm} \vspace{1mm}
      \textbf{Forming part of the city wall, the gate was intended by the more pragmatic Prussians not so much as a triumphal arch as an imposing tollgate for collecting duties.
      }  \end{minipage} & \begin{minipage}{4cm} \vspace{8mm} \textbf{The gate was just a triumphal arch 
      \\
      \\
      \\}  \end{minipage} \\ 
        \bottomrule
    \end{tabular} }
    \vspace{1.5mm}
    \caption{Four examples showing the Arabic text data, English translation and label (0: entailment, 1: neutral, 2:contradiction).}
    \label{beispiel_korpus}
\end{table*}

As the data was compiled from different sources, necessary standard normalizations with respect to encoding, column names, label mappings etc. were performed.
 
 
In the context of using transformer-based models for NLI/CD tasks, additional preprocessing steps such as stemming or stopword removal are not necessary. 

After performing the general preprocessing for the data from the three sources, the next step is to merge the data to create a unified dataset. To ensure a fair distribution of training, testing, and validation data, the merged data is being randomly shuffled before splitting. The split is then done with a distribution of 60\% for training data, 20\% for testing data, and 20\% for validation data.

The final data set consists of a total of 14,758 pairs of premises and hypotheses, each accompanied by an English translation and a label. The labels are encoded as 0 (entailment), 1 (neutral), and 2 (contradiction). Retaining the English translation of the data can assist in better comparing and understanding different texts and contexts, facilitating the comparison of various models.

For the NER pre-training, the ANERcorp corpus from CamelLabSplits \cite{CamelLab} is being used, which contains 3973 text samples in Arabic language with annotations for the NER task. The entity types "person", "location", "organization" and "miscellaneous" are being used and annotated according to the IOB-scheme.

\section*{Methodology}
In the scope of this study, analyses are conducted for Natural Language Inference (NLI) and Contradiction Detection (CD). The data processing pipeline remains the same for both tasks since the input data consists of the "premise" and "hypothesis" columns in both cases. However, differences arise in terms of the labels used. The labels of the datasets for the CD task are modified to treat the problem as a binary classification. Originally, "0" was used for entailment, "1" for neutral, and "2" for contradiction. However, since only a binary outcome is required in case of CD, "0" and "1" are mapped to "0", meaning there is no contradiction, and "2" to "1", meaning there is a contradiction.

In this study, two state-of-the-art models, namely AraBERT \cite{AraBERT} and XLM-RoBERTa \cite{XLM-RoBERTa}, are being investigated. AraBERT is based on the BERT \cite{BERT} paradigm and pre-trained on 24GB of news corpora in Arabic language with the Masked Language Modeling (MLM) and Next Sentence Prediction (NSP) objectives. A special sub-word segmentation algorithm is used to account for the semantic granularity of the Arabic language. For this study, the base version of AraBERT with 136M parameters is being used. XLM-RoBERTa is a multi-lingual language model, which has been pre-trained for MLM on 2.5TB of CommonCrawl data in 100 languages. It is shown to achieve state-of-the-art results on many NLP tasks across multiple languages. We use the base version with 279M parameters. To this end, we conduct the following modeling and evaluation steps:

\begin{itemize}

 \item    Loading the pre-trained model: 
 We utilize the checkpoints "xlm-roberta-base"\footnote{\url{https://huggingface.co/xlm-roberta-base}} and "aubmindlab/bert-base-arabertv02"\footnote{\url{https://huggingface.co/aubmindlab/bert-base-arabertv02}} for XLMRoBERTa and AraBERT, respectively.

 \item Additional pre-training: We employ Named Entity Recognition (NER) as an additional pre-training step before fine-tuning the model on the downstream task. Following the findings from \cite{M. Pielka}, the idea is to provide the transformer with semantic knowledge that would help in identifying contradictions and entailments. In the pre-training phase, a word-level classification head is being attached to the encoder and later discarded. 

  \item   Model fine-tuning: The pre-trained model is then fine-tuned on the tasks (NLI and CD) using the created datasets.

 \item    Hyperparameter optimization is performed during the finetuning process to enhance the model's performance. Techniques such as learning rate scheduling, dropout, and batch size adjustment are employed. 

\item  Model evaluation: After model fine-tuning, the performance of the model is evaluated using metrics such as accuracy or F1 score. 
    
\end{itemize}
The final preparation of input data involves merging the "premise" and "hypothesis" columns in the training, testing, and validation datasets using special tokens ([CLS] and [SEP]). The separator token "[SEP]" is recognized by the BERT tokenizer as a special token to separate different parts of the text. This allows the BERT model to process the meaning of each text part separately and then combine them in a single step. The classification token "[CLS]" in the BERT architecture signals the model to perform classification on the input text. It is placed at the beginning of the text and serves as an indication for the model to make predictions about the text's class membership.

Overall, this study demonstrates the effective utilization of transformer-based models, showcasing their adaptability and performance in NLI and CD tasks in the Arabic language. 

\section{Experiments and Results}
We mainly compare AraBERT and XLM-RoBERTa with respect to their performance on the NLI and CD task. An extensive hyperparameter search is being conducted, using the Optuna \cite{Optuna} tool with 150 optimization runs. We train the models for a maximum of 5 epochs, with the option of early stopping if the validation performance does not improve any further. The AdamW optimizer \cite{AdamW} is being applied. The other hyperparameters, including learning rate, weight decay and batch size, are chosen according to the best result from the respective Optuna run. The experimental results on the two tasks are displayed in tables \ref{nli_results} and \ref{ergebnisse Transformer_CD}.

\begin{table}[ht]
    \centering
    \begin{tabular}{lccc}
             \textbf{Model} & \textbf{Multitask Finetuning}  &   \textbf{Accuracy}   & \textbf{F1-Score} \\ \midrule

AraBERT & 
           x
          & 
75.3   & 
75.4  
         
          \\ \midrule
XLM-R  & 
           x
          & 
78.7   & 
          78.8   
         
         \\ \midrule
AraBERT & 
           \checkmark
          & 
76.8    & 76.8   
         
          \\ \midrule
XLM-R & 
           \checkmark
          & 
 \textbf{78.9}    & 
         \textbf{79.0}  
         
         \\

        \bottomrule
    \end{tabular} 
  \vspace{1.5mm}
    \caption{Results for the NLI task with AraBERT \& XLM-RoBERTa, in \%. Accuracy and macro average F1-score are being reported.}
    \label{nli_results}
\end{table}

\begin{table}[ht]
    \centering
    \begin{tabular}{lccc}
             \textbf{Model} & \textbf{Multitask Finetuning} & \textbf{Accuracy}   & \textbf{F1-Score}  \\ \midrule
       
AraBERT & 
           x
          & 87.4  & 
82.3  \\ \midrule
       
XLM-R & 
          x
          & 
86.9 & 
         81.0  \\ \midrule

AraBERT & 
           \checkmark
          & 
 \textbf{88.1}  & 
    \textbf{84.5}    \\ \midrule
         
     XLM-R& 
         \checkmark
          & 
86.8   & 
          81.1  \\ \midrule
    \end{tabular} 
  \vspace{1.5mm}
    \caption{Results for the CD task with AraBERT \& XLM-RoBERTa, in \%. Accuracy and macro average F1-score are being reported.}
    \label{ergebnisse Transformer_CD}
\end{table}

All the models achieve overall good results on both tasks. It is especially noteworthy that AraBERT performs competitively with XLM-RoBERTa, even though it was pre-trained on a considerably smaller amount of data. With respect to the CD task, the best AraBERT model with mutlitask finetuning even outperforms XLM-RoBERTa by two percentage points. This emphasizes the fact that language-specific finetuning can be more effective than extensive multi-lingual pre-training for some downstream tasks. We also find that adding the NER objective as an additional pre-training step improves the performance. Interestingly, this effect is stronger for the smaller AraBERT model, suggesting that it can help bridge the performance gap that is caused by XLMRoBERTa's larger model size and the amount of training data it has access to.

\section{Conclusion and Outlook}
We presented the first comprehensive study on Natural Language Inference and Contradiction Detection in Arabic language, in which we applied state-of-the-art transformer methods combined with an informed pre-training approach. The methods achieve promising results on our custom data set, emphasizing the fact that smaller, language-specific models like AraBERT can perform competitively with larger multi-lingual models, if they are being enhanced with additional linguistic knowledge. Further, we collected a large data set for NLI in Arabic language.

Future work includes adding more pre-training methods such as Part of Speech tagging, Word Sense Disambiguation or Semantic Role Labeling. We expect the performance of the AraBERT model to improve even further by adding more linguistic knowledge. Another direction of research is to exploit the potential of large language models such as GPT-4 \cite{GPT4}, by casting the classification problem as a text generation task. It would be interesting to see the performance of those resourceful models when confronted with a low-resource language such as Arabic.

\section*{Acknowledgements}
This research has been partially funded by the Federal Ministry of Education and Research of Germany and the state of North-Rhine Westphalia as part of the Lamarr-Institute for Machine Learning and Artificial Intelligence.

\end{document}